\def\year{2022}\relax
\documentclass[letterpaper]{article} 
\usepackage{aaai22}  
\usepackage{times}  
\usepackage{helvet}  
\usepackage{courier}  
\usepackage[hyphens]{url}  
\usepackage{graphicx} 
\usepackage{booktabs} 
\usepackage{natbib}  
\usepackage{caption} 
\DeclareCaptionStyle{ruled}{labelfont=normalfont,labelsep=colon,strut=off} 
\usepackage[utf8]{inputenc} 
\usepackage[T1]{fontenc}    
\usepackage{hyperref}       
\usepackage{amsfonts}       
\usepackage{nicefrac}       
\usepackage{microtype}      
\usepackage{algpseudocode}
\usepackage[ruled,vlined]{algorithm2e}
\usepackage{amsmath,bm}
\usepackage{amsthm}
\usepackage{subfigure}
\usepackage{color}
\usepackage{multirow}
\usepackage{balance}
\usepackage{enumitem}
\usepackage{wrapfig}
\usepackage[english]{babel}

\usepackage{xcolor}         
\algnewcommand{\LeftComment}[1]{ \(\triangleright\) #1}

\title{Meta Propagation Networks for Graph Few-shot Semi-supervised Learning}
\author {
    Anonymous Author(s)
}

\author {
    Kaize Ding$^\dagger$,
    Jianling Wang$^\ddagger$,
    James Caverlee$^\ddagger$,
    \textrm{and} Huan Liu$^\dagger$
}
\affiliations {
    $^\dagger$Arizona State University\\
    $^\ddagger$Texas A\&M University\\
    $^\dagger$\texttt{\{kaize.ding, huan.liu\}@asu.edu} \\
    $^\ddagger$\texttt{\{jlwang, caverlee\}@tamu.edu}
}

\begin{document}

\maketitle

\begin{abstract}
 Inspired by the extensive success of deep learning, graph neural networks (GNNs) have been proposed to learn expressive node representations and demonstrated promising performance in various graph learning tasks. However, existing endeavors predominately focus on the conventional semi-supervised setting where relatively abundant gold-labeled nodes are provided. While it is often impractical due to the fact that data labeling is unbearably laborious and requires intensive domain knowledge, especially when considering the heterogeneity of graph-structured data. Under the few-shot semi-supervised setting, the performance of most of the existing GNNs is inevitably undermined by the overfitting and oversmoothing issues, largely owing to the shortage of labeled data. In this paper, we propose a decoupled network architecture equipped with a novel meta-learning algorithm to solve this problem. In essence, our framework Meta-PN infers high-quality pseudo labels on unlabeled nodes via a meta-learned label propagation strategy, which effectively augments the scarce labeled data while enabling large receptive fields during training. Extensive experiments demonstrate that our approach offers easy and substantial performance gains compared to existing techniques on various benchmark datasets. The implementation of Meta-PN is publicly available at \url{https://github.com/kaize0409/Meta-PN}.



\end{abstract}

\section{Introduction}

Graphs serve as a common language for modeling a plethora of structured and relational systems, ranging from social networks~\cite{zafarani2014social} to citation networks~\cite{namata2012query}, to molecular graphs~\cite{klicpera2019directional}. To ingest the rich information encoded in graph-structured data, it is of paramount importance to learn expressive node representations by modeling the information from both node attributes and graph topology. Among numerous endeavors in the graph machine learning (Graph ML) community, graph neural networks (GNNs) have received significant attention due to their effectiveness and scalability~\cite{kipf2017semi,velickovic2018graph,hamilton2017inductive}.

In general, most of the prevailing GNNs adopt the message-passing scheme to learn the representation of a node by iteratively transforming, and propagating/aggregating node features from its local neighborhoods. Along with this idea, different designs of GNN architectures have been proposed, including graph convolutional networks (GCNs)~\cite{kipf2017semi,defferrard2016convolutional}, graph attention networks (GAT)~\cite{velickovic2018graph,wang2019heterogeneous} and many others~\cite{hamilton2017inductive,xu2019powerful,klicpera2019predict,wu2019simplifying,chen2020simple}. Despite their promising results, existing GNNs developed for \textit{semi-supervised} node classification predominantly assume that the provided gold-labeled nodes are relatively abundant. This assumption is often impractical as data labeling requires intensive domain knowledge, especially when considering the heterogeneity of graph-structured data~\cite{yao2020graph,ding2020graph}. When only few labeled nodes per class are available, how to improve the expressive power of Graph ML models for tackling the \textit{few-shot semi-supervised} node classification problem remains understudied and meanwhile requires urgent research efforts. 






However, it is a non-trivial and challenging task mainly because of two reasons: (i) \textit{\textbf{oversmoothing and overfitting.}} In general, most of the existing GNNs are designed with shallow architecture with restricted receptive fields, thereby restricting the efficient propagation of label information~\cite{li2018deeper}. In order to propagate the label signals more broadly, larger receptive fields of GNNs, i.e., the number of layers, are particularly desirable~\cite{klicpera2019predict}. Due to the entanglement of representation transformation and propagation in each layer, GNNs will face the oversmoothing issue when increasing the model depth~\cite{liu2020towards}, which in turn renders the learned node representations inseparable. In the meantime, when training with few labeled nodes, an over-parametric deep GNN model tends to overfit and goes timber easily; (ii) \textit{\textbf{no auxiliary knowledge.}} Though previous works proposed for graph few-shot learning~\cite{ding2020graph} or cross-network transfer learning~\cite{yao2020graph} also focus on related low-resource scenarios, their key enabler lies in transferring knowledge from either label-rich node classes or other similar networks. Nonetheless, such auxiliary knowledge is commonly not accessible, making those methods practically infeasible to be applied to few-shot semi-supervised learning. As suggested by previous research, pseudo-labeling~\cite{li2018deeper,sun2020multi,ding2022data} is commonly beneficial to solve semi-supervised learning, whereas inaccurate pseudo labels may instead lead to abysmal failure. Hence, how to infer accurate pseudo labels on unlabeled nodes plays a pivotal role to solve the studied research problem.

To address the aforementioned challenges, we propose a new graph meta-learning framework, Meta Propagation Networks (Meta-PN), which goes beyond the canonical message-passing scheme of GNNs and learns expressive node representations in a more label-efficient way. Specifically, Meta-PN is built with two simple neural networks, i.e., \textit{adaptive label propagator} and \textit{feature-label transformer}, which inherently decouples the entangled propagation and transformation steps of GNNs, thereby allowing sufficient propagation of label signals without suffering the oversmoothing issue. At its core, the \textit{adaptive label propagator} is meta-learned to adjust its propagation strategy for inferring accurate pseudo labels on unlabeled nodes, according to the feedback (i.e., the performance change on the gold-labeled nodes) from the target model \textit{feature-label transformer}. This way the generated soft pseudo labels not only capture informative local and global structure information, but more importantly, have aligned data usage with the gold-labeled nodes. Optimizing with our proposed meta-learning algorithm, those two decoupled networks are able to reinforce each other synergistically. As a result, the target model assimilates the encoded knowledge of pseudo-labeled nodes and offers excellent performance for the semi-supervised node classification problem even if only few labeled nodes are available. In summary, the contributions of our work are as follows: 









\begin{itemize}
\item We study the problem of semi-supervised node classification under the few-shot setting, which remains largely under-studied in the Graph ML community.


\item We propose a simple yet effective graph meta-learning framework Meta-PN to solve the studied problem. The essential idea is to augment the limited labeled data via a meta-learned label propagation strategy.


\item We conduct comprehensive evaluations on different graph benchmark datasets to corroborate the effectiveness of Meta-PN. The results show its superiority over the state-of-the-arts on semi-supervised node classification, especially under the low-resource setting.
\end{itemize}

\section{Related Work}

\noindent\textbf{Graph Neural Networks.} Graph neural networks (GNNs), a family of neural models for learning latent node representations in a graph, have achieved gratifying success in different graph learning tasks~\cite{defferrard2016convolutional,kipf2017semi}. Originally inspired by graph spectral theory, spectral-based graph convolutional networks (GCNs)~\cite{defferrard2016convolutional,kipf2017semi,wu2019simplifying} extend the convolution operation in the spectral domain to network representation learning. Among them, the model proposed by Kipf et al.~\cite{kipf2017semi} has become the most prevailing one by using a linear filter. Afterwards, spatial-based graph neural networks that follow the message-passing scheme have been extensively investigated~\cite{hamilton2017inductive,velickovic2018graph,xu2019powerful}. Those methods follow the homophily principle~\cite{mcpherson2001birds} and learn node representations by iteratively transforming, and propagating/aggregating node features within graph neighborhoods. For example, GAT~\cite{velickovic2018graph} and GraphSAGE~\cite{hamilton2017inductive} adopt different strategies to specify fine-grained weights on neighbors when aggregating neighborhood information of a node.




\smallskip
\noindent\textbf{Deep Graph Neural Networks.} Despite the success of GNNs, the notorious over-smoothing issue can largely undermine the model performance when increasing the model depth. To counter this, researchers also try to increase the message-passing range or receptive fields of GNNs by proposing different techniques, such as adding advanced normalizations~\cite{zhao2019pairnorm,li2019deepgcns}, decoupling the feature transformation and propagation steps~\cite{wu2019simplifying,klicpera2019predict,liu2020towards,dong2021equivalence} and many others~\cite{li2018deeper,xu2018representation}. In particular, decoupled graph neural networks have become a prevailing paradigm in the community due to their simplicity and learning efficiency~\cite{klicpera2019predict,dong2021equivalence,chien2021adaptive}. For example, APPNP~\cite{klicpera2019predict} propagates the neural predictions via personalized PageRank, which can preserve the node’s local information while increasing the receptive fields. DAGNN~\cite{liu2020towards} decouples the propagation and transformation steps and then utilizes an adaptive adjustment mechanism to balance the information from local and global neighborhoods of each node. However, these deep GNNs are not specifically developed to tackle the low-resource settings, especially when only very few labels are available. 




\smallskip
\noindent\textbf{Graph Learning with Few Labels.}
For real-world graph learning tasks, the amount of gold-labeled samples is usually quite limited due to the expensive labeling cost. To improve the GNN model performance on the node classes with only few labeled nodes, graph few-shot learning~\cite{zhou2019meta,ding2020graph,wang2020graph} and cross-network transfer learning~\cite{yao2020graph,ding2021few} have been proposed to transfer the knowledge from other auxiliary data source(s).
Nonetheless, for the problem of \textit{few-shot semi-supervised} node classification, such auxiliary datasets are commonly not allowed to use. As another line of related work, Li et al.~\cite{li2018deeper} combined GCNs and self-training to expand supervision signals, while M3S~\cite{sun2020multi} advances this idea by utilizing the clustering method to eliminate the inaccurate pseudo labels. However, those methods cannot directly address the oversmoothing issue and may suffer from inaccurate pseudo labels. By conducting meta-learning on top of a decoupled design, our approach Meta-PN achieves superior performance on few-shot semi-supervised node classification.

\section{Proposed Approach}

 \begin{figure*}[!ht]
    \graphicspath{{figures/}}
    \centering
    \includegraphics[width=0.92\textwidth]{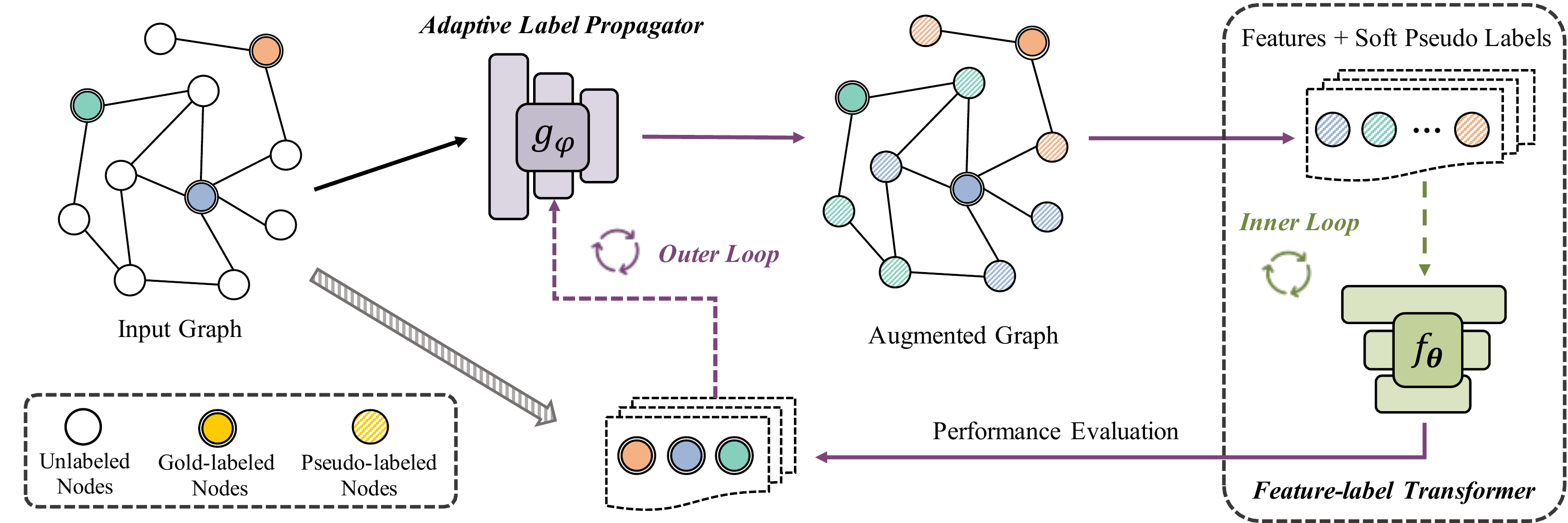}
    \caption{Illustration of our Meta-PN framework. The \textit{adaptive label propagator} propagates known labels to unlabeled nodes and the \textit{feature-label transformer} transforms the features of each node to a soft label vector. Specifically, the \textit{adaptive label propagator} is meta-learned to adjust its label propagation strategy to infer accurate pseudo labels on unlabeled nodes, according to the \textit{feature-label transformer}'s performance change on the labeled nodes. We omit the node features for simplicity.}%
    \label{fig:framework}%
\end{figure*}

We first introduce the notations used throughout this paper. Let $G = (\mathcal{V}, \mathcal{E})$ denote an undirected graph with nodes $\mathcal{V}$ and edges $\mathcal{E}$. $\mathcal{V}^L$ and $\mathcal{V}^U$ stand for labeled and unlabeled node set, respectively. Let $n$ denote the number of nodes and $m$ the number of edges. The nodes in $G$ are described by the attribute matrix $\mathbf{X} \in \mathbb{R}^{n \times f}$, where $f$ denotes the number of features per node. The graph structure of $G$ is described by the adjacent matrix $\mathbf{A} \in \{0, 1\}^{n \times n}$, while $\Tilde{\mathbf{A}}$ stands for the adjacency matrix for a graph with added self-loops. We let $\mathbf{D}$ and $\Tilde{\mathbf{D}}$ be the diagonal degree matrix of $\mathbf{A}$ and $\Tilde{\mathbf{A}}$, respectively. Moreover, $\Tilde{\mathbf{A}}_{sym} = \Tilde{\mathbf{D}}^{-\frac{1}{2}} \Tilde{\mathbf{A}}\Tilde{\mathbf{D}}^{-\frac{1}{2}}$ denote the symmetric normalized adjacency matrix with self-loops. The class (or label) matrix is represented by $\mathbf{Y} \in \mathbb{R}^{n \times c}$, where $c$ denotes the number of classes.

\subsection{Architecture Overview} 
For solving the problem of few-shot semi-supervised node classification, we propose a new framework Meta Label Propagation (Meta-PN), which is built with two simple neural networks, i.e., \textit{adaptive label propagator} and \textit{feature-label transformer}. By decoupling the propagation and transformation steps with two independent networks, such a design inherently allows large receptive fields without suffering performance deterioration. Upon our proposed meta-learning algorithm, the meta learner -- \textit{adaptive label propagator} learns to adjust its propagation strategy for inferring accurate pseudo labels on unlabeled nodes, by using the feedback from the target model. Meanwhile, the target model -- \textit{feature-label transformer} assimilates both the structure and feature knowledge from pseudo-labeled nodes, therefore addressing the challenges behind few-shot semi-supervised learning. Specifically, we introduce the architecture details as follows:

\paragraph{Adaptive Label Propagator (Meta Learner).} In order to enable broader propagation of label signals, we propose to adopt the idea of label propagation (LP)~\cite{zhu2002learning} to encode informative local and global structural information. Similar to the message-passing scheme adopted by many GNNs, label propagation follows the principle of Homophily~\cite{mcpherson2001birds} that indicates two connected nodes tend to be similar (share same labels). Specifically, the objective of LP is to find a prediction matrix $\mathbf{\hat{Y}} \in \mathbb{R}^{n \times c}$ that agrees with the label matrix $\mathbf{Y}$ while being smooth on the graph such that nearby vertices have similar soft labels~\cite{zhou2004learning}. Generally, the solution can be approximated via the iteration as follows:
\begin{equation}
\label{eq:lp}
   \mathbf{\hat{Y}} = \mathbf{Y}^{(K)},  \mathbf{Y}^{(k+1)} = \mathbf{T} \mathbf{Y}^{(k)}, 
\end{equation}
where $\mathbf{Y}^{(0)} = \mathbf{Y}$ and $K$ denotes the number of power iteration (propagation) steps. The transition matrix is denoted by $\mathbf{T}$, which can be set as any form of normalized adjacency matrix (e.g., $\Tilde{\mathbf{A}}_{sym}$). After $K$ iterations of label propagation, the predicted soft label matrix $\mathbf{\hat{Y}}$ can capture the prior knowledge of neighborhood label distribution up to $K$ hops away.

In practice, various propagation schemes can be adopted for LP, such as the Personalized PageRank~\cite{klicpera2019predict} where $\mathbf{Y}^{(k+1)} = (1 - \alpha) \mathbf{T} \mathbf{Y}^{(k)} + \alpha \mathbf{Y}^{(0)}$. With appropriate teleport probability $\alpha$, the smoothed labels can avoid losing the focus on local neighborhood even using infinitely many propagation steps~\cite{klicpera2019predict}. However, most of the existing LP algorithms cannot adaptively balance the label information from different neighborhoods for each node, which largely restricts the model expressive power when learning with complex real-world graphs.






To counter this issue, we build an \textit{adaptive label propagator} $g_{\bm\phi}(\cdot)$ parameterized with $\bm\phi$, which is able to adjust the contribution of different propagation steps for computing the smoothed label vector of one node. Specifically, the propagation strategy can be formulated as:
\begin{equation}
\mathbf{\hat{Y}}_{i,:} = \sum_{k=0}^{K} \gamma_{ik} \mathbf{Y}^{(k)}_{i,:},  \mathbf{Y}^{(k+1)} = \mathbf{T} \mathbf{Y}^{(k)}, 
\end{equation}
where $\gamma_{ik}$ denotes the influence from $k$-hop neighborhood for node $v_i$ and can be computed by the attention mechanism:
\begin{equation}
\gamma_{ik} = \frac{\exp \Big( \mathbf{a}^{\mathrm{T}} \text{ReLU} \Big(\mathbf{W} \mathbf{Y}_{i,:}^{(k)}\Big)\Big)}{\sum_{k'=0}^{K}\exp \Big(\mathbf{a}^{\mathrm{T}} \text{ReLU}\Big(\mathbf{W} \mathbf{Y}_{i,:}^{(k')}\Big)\Big)},
\end{equation}
where $\mathbf{a} \in \mathbb{R}^c$ is the attention vector and $\mathbf{W} \in \mathbb{R}^{c \times c}$ is a weight matrix. By setting the attention vector and weight matrix as learnable parameters, the \textit{adaptive label propagator} acquire the capability of  adjusting its propagation strategy for each node and the final smoothed labels can capture rich structure information of the input graph.

\paragraph{Feature-label Transformer (Target Model).} After encoding the structure knowledge into the smoothed label matrix $\mathbf{\hat{Y}}$, we then build a \textit{feature-label transformer} $f_{\bm\theta}(\cdot)$ that transforms node features to node label, in order to further capture feature-based graph information. For each node $v_i$, the \textit{feature-label transformer} parameterized with $\bm\theta$ takes the node feature vector $\mathbf{X}_{i,:}$ as input and predicts its node label $\mathbf{P}_{i,:}$ by:
\begin{equation}
    \mathbf{P}_{i,:} =  f_{\bm\theta}(\mathbf{X}_{i,:}),
\end{equation}
where $f_{\bm\theta}(\cdot)$ is a multi-layer perceptron (MLP) followed by a softmax function. 

In order to learn the target model , i.e., \textit{feature-label transformer}, we take the soft pseudo labels computed by the \textit{adaptive label propagator} as ``ground-truth''. Ideally, if the generated pseudo labels are of high quality, they can be used to augment the insufficient labeled nodes to avoid overfitting and improve the model generalization ability~\cite{li2018deeper}. In the meantime, high-quality pseudo-labeled data not only encodes the feature patterns of unlabeled nodes, but also carries informative local and global structure knowledge, which enables the target model to leverage larger receptive fields without suffering from performance degradation. As a result, the \textit{feature-label transformer} can achieve excellent performance on the problem of few-shot semi-supervised node classification.


It is worth mentioning that, the target model trained with meaningful pseudo labels can be considered as a special variant of GCN, which allows far more propagation steps with much fewer parameters. Due to the space limit, we attach the detailed proof in Appendix \ref{proof}.





\paragraph{Learning to Propagate.} One key challenge of our approach lies in how to learn a better label propagation strategy for generating pseudo labels on unlabeled nodes. If the pseudo labels are inaccurate, the target model may easily overfit to mislabeled nodes and encounter severe performance degradation~\cite{ren2018learning}. This issue is also known as the problem of confirmation bias in pseudo-labeling~\cite{arazo2020pseudo}. While inferring accurate pseudo labels by recursively selecting a subset of samples, re-training the prediction model will be too expensive and unstable. Hence, without linking the two networks in a principled way, it is almost infeasible to enforce the \textit{adaptive label propagator} to infer meaningful label propagation strategy for improving the performance of the \textit{feature-label transformer}.


In this work, we propose to tackle this problem through a unified meta-learning algorithm, allowing the model to infer accurate pseudo labels for unlabeled nodes and learn a better target model. In a sense, if the generated pseudo labels are of high quality, their data utility should align with the gold-labeled nodes. Accordingly, we can derive the following meta-learning objective: \textit{optimal pseudo labels generated by meta-learner should maximize target model's performance (minimize the classification loss) on the gold-labeled training nodes}. For each \textit{meta label propagation} task,
the goal is to generate pseudo labels for a batch of unlabeled nodes using the feedback of the target model (i.e., \textit{feature-label transformer}). By optimizing the \textit{adaptive label propagator} on a meta-level, it can adjust the label propagation strategy to generate informative pseudo-labeled data.





\subsection{Model Learning via Bi-level Optimization}

The above meta-learning objective implies a bi-level optimization problem with $\bm\phi$ as the outer-loop parameters and $\bm\theta$ as the inner-loop parameters. This problem shares the same formulation with many meta-learning algorithms that have been proposed for solving different learning tasks such as few-shot learning~\cite{finn2017model}, hyper-parameter optimization~\cite{baydin2018online}, and neural architecture search~\cite{liu2018darts}. Specifically, let $\mathcal{L}$ denote the cross-entropy loss for node classification, and this bi-level optimization problem can be formulated as:
\begin{equation}
\begin{aligned}
&\text{Outer loop: }\\
& \bm\phi^* = \arg \min_{\bm\phi} \mathbb{E}_{v_i \in \mathcal{V}^L} [ \mathcal{L} (f_{\bm\theta^*(\bm\phi)}(\mathbf{X}_{i,:}), \mathbf{Y}_{i,:})],\\
&\text{Inner loop: }\\
& \bm\theta^*(\bm\phi) = 
\arg \min_{\bm\theta} \mathbb{E}_{v_i \in \mathcal{V}^U} [ \mathcal{L}(f_{\bm\theta}(\mathbf{X}_{i,:}), g_{\bm\phi}(\mathbf{Y}, \mathbf{A})_{i,:})].
\label{eq:bi-level}
\end{aligned}
\end{equation}

The optimal solution of this bi-level optimization problem can potentially train a highly discriminative \textit{feature-label transformer} with abundant pseudo-labeled data and only a small set of gold-labeled data. However, deriving exact solutions for this bi-level problem is indeed analytically intractable and computationally expensive, owing to the fact that it requires solving for the optimal $\bm\theta^*(\bm\phi)$ whenever $\bm\phi$ gets updated. To approximate the optimal solution $\bm\theta^*(\bm\phi)$, we propose to take one step of gradient descent update for $\bm\theta$, without solving the inner-loop optimization completely by training until convergence. This way allows the optimization algorithm to alternatively update the parameters of \textit{feature-label transformer} in the inner loop and the parameters of \textit{adaptive label propagator} in the outer loop:





\paragraph{Target Model (Inner-loop) Update.} Given a batch of unlabeled nodes from $\mathcal{V}^U$, we update the target model parameters $\bm\theta$ by taking their pseudo labels computed by 
the \textit{adaptive label propagator} as ground-truth. For simplicity, we use $J_{\text{pseudo}}(\bm\theta, \bm\phi)$ to denote the inner-loop loss computed on a batch of pseudo-labeled nodes. Assuming that parameter $\bm\theta$ is updated using the computed gradient descent on $J_{\text{pseudo}}(\bm\theta, \bm\phi)$, with a learning rate $\eta_{\bm\theta}$, then we have:
\begin{equation}
    \bm\theta' = \bm\theta - \eta_{\bm\theta} \nabla_{\bm\theta} J_{\text{pseudo}}(\bm\theta, \bm\phi).
    \label{eq:inner}
\end{equation}



\paragraph{Meta Learner (Outer-loop) Update.} Note that the dependency between $\bm\phi$ and $\bm\theta$ allows us to compute the meta-level (outer-loop) loss using the gold-labeled nodes from $\mathcal{V}^L$. We denote this loss by $J_{\text{gold}}( \bm\theta'(\bm\phi))$ for the purpose of simplicity, and back-propagate this loss to compute the gradient for the \textit{feature-label transformer}. Having the gradient, we can update on the backward parameters $\bm\phi$ with learning rate $\eta_{\bm\phi}$:
\begin{equation}
    \bm\phi' = \bm\phi - \eta_{\bm\phi} \nabla_{\bm\phi} J_{\text{gold}}(\bm\theta'(\bm\phi)).
    \label{eq:outer}
\end{equation}

To further compute the gradient of $\bm\phi$, we apply chain rule to differentiate $J_{\text{gold}}( \bm\theta'(\bm\phi))$ with respect to $\bm\phi$ via $\bm\theta'$, where $ \bm\theta'(\bm\phi) = \bm\theta - \eta_{\bm\theta} \nabla_{\bm\theta} J_{\text{pseudo}}(\bm\theta, \bm\phi)$. The full derivation is delegated to the Appendix \ref{gradient}. Here, we directly present the final result:
\begin{equation}
\begin{aligned}
     \nabla_{\bm\phi} J_{\text{gold}}(\bm\theta'(\bm\phi)) \approx - \frac{\eta_{\bm\phi}}{2\epsilon}&[\nabla_{\bm\phi} J_{\text{pseudo}}(\bm\theta^+, \bm\phi) -\\
     & \nabla_{\bm\phi} J_{\text{pseudo}}(\bm\theta^-, \bm\phi)],
    \label{eq:gradient}
\end{aligned}
\end{equation}
where $\bm\theta^{\pm} = \bm\theta \pm \epsilon \nabla_{\bm\theta'} J_{\text{gold}}(\bm\theta'(\bm\phi))$, and $\epsilon$ is a small scalar for
finite difference approximation.


By alternating the update rules
in Eq. (\ref{eq:inner}) and Eq. (\ref{eq:outer}), we are able to progressively learn the two modules. The complete meta-learning algorithm is shown in Algorithm \ref{alg:Meta-PN}. 
Finally, as the \textit{feature-label transformer} only learns from unlabeled data with pseudo labels generated by the
\textit{adaptive label propagator}, we can further fine-tune the \textit{feature-label transformer} on labeled data to improve its accuracy. After the model converges, we use the \textit{feature-label transformer} to make final predictions on unlabeled nodes. 

\begin{algorithm}[t!]
\caption{The learning algorithm of Meta-PN.}
\label{alg:Meta-PN}
\LinesNumbered
\small
\KwIn{The input graph $G = (\mathcal{V}, \mathcal{E})$ with labeled node set $\mathcal{V}^L$ and unlabeled node set $\mathcal{V}^U$; Batch size $B$}
\KwOut{The well-trained \textit{feature-label transformer}}

Initialize the parameters $\bm\theta$ and $\bm\phi$



\While{not converge}{
    
Randomly sample a batch of $B$ unlabeled nodes

\LeftComment{\textit{Pseudo Label Generation}}

Compute the pseudo labels for sampled nodes using the \textit{adaptive label propagator} $g_{\bm\phi} (\cdot)$

\LeftComment{\textit{Inner-loop Update for $\bm\theta$}}

Compute $J_{\text{pseudo}}(\bm\theta, \bm\phi)$ using the generated pseudo labeled nodes 

Update parameters $\bm\theta$ of the \textit{feature-label transformer} $f_{\bm\theta}(\cdot)$ according to Eq. (\ref{eq:inner}) 

\LeftComment{\textit{Outer-loop Update for $\bm\phi$}}

Compute $J_{\text{gold}}( \bm\theta'(\bm\phi))$ on $\mathcal{V}^L$ using the updated \textit{feature-label transformer} 

Update parameters $\bm\phi$ of the \textit{adaptive label propagator}  $g_{\bm\phi} (\cdot)$ according to Eq. (\ref{eq:outer}) and Eq. (\ref{eq:gradient})

}

Fine-tune the \textit{feature-label transformer} using $\mathcal{V}^L$.
    
\Return The well-trained \textit{feature-label transformer} 

\end{algorithm}




\section{Experiments}

We evaluate the effectiveness of our approach on different benchmark datasets. Specifically, our evaluation centers around three questions: (i) can Meta-PN outperform state-of-the-art GNN models when labeled data is extremely sparse? (ii) compared with the state-of-the-art GNNs, can Meta-PN achieve competitive performance under the standard semi-supervised setting? and (iii) when the data-scale goes large, how would Meta-PN perform compared to other methods?

\subsection{Experimental Setup}
\label{sec:experiment}
\smallskip
\noindent \textbf{Evaluation Datasets.} 
We conduct experiments on five graph benchmark datasets for semi-supervised node classification to demonstrate the effectiveness of the proposed Meta-PN. The detailed statistics of the datasets are summarized in Table \ref{table:datasets}. Specifically, \textbf{Cora-ML}, \textbf{CiteSeer}~\cite{sen2008collective} and \textbf{PubMed} \cite{namata2012query} are the three most widely used citation networks. \textbf{MS-CS} is a co-authorship network based on the Microsoft Academic Graph~\cite{shchur2018pitfalls}. For data splitting, we follow the previous work~\cite{klicpera2019predict} and split each dataset into training set (i.e., K nodes per class for K-shot task), validation set and test set. In addition, to further evaluate the performance of different methods on large-scale graphs, we further include the \textbf{ogbn-arxiv} datasets from Open Graph Benchmark (OGB)~\cite{hu2020open}. For the ogbn-arxiv dataset, we randomly sample 1.0\%, 1.5\%, 2.0\%, 2.5\% nodes from its training splits as labeled data while using the same validation and test splits in OGB Benchmark~\cite{hu2020open}. Note that for all the datasets, we run each experiment 100 times with multiple random splits and different initializations. 




\begin{table}[t!]
\caption{Summary statistics of the evaluation datasets.}
\centering
\scalebox{0.9}{
\begin{tabular}{@{}lccccccc@{}}
\toprule


\textbf{Dataset} & \# Nodes & \# Edges & \# Features & \# Classes \\ \midrule
Cora-ML & 2,810 & 7,981 &  2,879 & 7 \\
CiteSeer & 2,110 & 3,668 & 3,703  & 6 \\
PubMed   & 19,717  &  44,324 &  500 & 3 \\
MS-CS  & 18,333   & 81,894 & 6,805 & 15 \\
ogbn-arxiv & 169,343  & 1,166,243 & 15 & 40 \\

\bottomrule
\end{tabular}}

\label{table:datasets}
\end{table}

\begin{table*}[t!]
\centering
\caption{Test accuracy on few-shot semi-supervised node classification: mean accuracy ($\%$) $\pm$ 95$\%$ confidence interval. }
\scalebox{0.825}{
\begin{tabular}{@{}lcccccccccccccccccc@{}}
\toprule
\rule{0pt}{10pt} \multirow{2}{*}{\textbf{Method}}  & \multicolumn{2}{c}{Cora-ML}  & &  \multicolumn{2}{c}{CiteSeer}  & & \multicolumn{2}{c}{PubMed} & & \multicolumn{2}{c}{MS-CS} \\ \cline{2-3} \cline{5-6} \cline{8-9} \cline{11-12} 

\rule{0pt}{10pt} & \multicolumn{1}{c}{3-shot}  & \multicolumn{1}{c}{5-shot}  & & \multicolumn{1}{c}{3-shot} &  \multicolumn{1}{c}{5-shot} & &
\multicolumn{1}{c}{3-shot} & \multicolumn{1}{c}{5-shot} & &
\multicolumn{1}{c}{3-shot} & \multicolumn{1}{c}{5-shot}

\\ \midrule

MLP       & $41.07\pm0.76$   & $51.12\pm 0.61$   & & $43.34\pm0.56$ & $44.90\pm0.60$ & & $56.59\pm0.93$ & $59.90\pm0.84$  &  & $70.33\pm0.37$& $79.41\pm0.31$ \\
LP        & $62.07\pm0.71$  & $68.01\pm0.62$  & & $54.07\pm0.59$ & $55.73\pm1.19$ &  & $58.75\pm0.89$ & $59.91\pm0.85$  &  & $57.96\pm0.69$& $62.98\pm0.61$ \\

GCN    & $48.02\pm0.89$  & $67.32\pm1.02$  & & $53.60\pm0.86$ & $62.60\pm0.58$ &  & $58.89\pm0.80$   &$65.77\pm0.98$  &  & $69.24\pm0.94$ & $84.43\pm0.89$ \\

SGC    & $49.60\pm0.55$  & $67.24\pm0.86$  & & $57.37\pm0.98$ & $61.55\pm0.53$ &  & $63.37\pm0.93$   &$64.93\pm0.81$  &  & $72.11\pm0.76$ & $87.51\pm0.27$   \\
\midrule

GLP   & $65.57\pm0.26$  & $71.26\pm0.31$  & & $65.76\pm0.49$ & $71.36\pm0.18$ &  & $65.34\pm0.54$  &$65.26\pm0.29$  & & $86.10\pm0.21$ & $86.94\pm0.23$   \\
IGCN   & $66.60\pm0.29$  & $72.50\pm0.20$  & & $67.47\pm0.29$ & \underline{$72.92\pm0.10$} &  & $62.28\pm0.23$  &$65.19\pm0.13$  &  & $85.83\pm0.06$ & $87.01\pm0.05$  \\
M3S  & $64.66\pm0.31$  & $69.64\pm0.18$  & & $65.12\pm0.20$ & $68.18\pm0.18$ &  & $63.40\pm0.32$  &$68.85\pm0.26$  &  & $84.96\pm0.18$ & $86.83\pm0.29$  \\
\midrule

APPNP   & \underline{$72.39\pm0.98$}  & \underline{$78.32\pm0.58$}  & & \underline{$67.55\pm0.77$} & $71.08\pm0.61$ &  & $70.52\pm0.62$  & \underline{$74.24\pm0.87$}  & & \underline{$86.65\pm0.42$} & $90.13\pm0.86$    \\
DAGNN   & $71.86\pm0.75$  & $77.20\pm0.69$  & & $66.62\pm0.27$ & $70.55\pm0.12$ &  & \underline{$71.22\pm0.82$}  & $73.91\pm0.71$  &  & $86.32\pm0.57$ & \underline{$90.30\pm0.66$}   \\
C\&S  & $68.93\pm0.68$  & $73.37\pm0.24$  & & $63.02\pm0.72$ & $64.72\pm0.53$ &  & $70.51\pm0.57$  & $73.22\pm0.57$  & & $85.86\pm0.45$ & $87.99\pm0.24$  \\
GPR-GNN   & $70.98\pm0.84$  & $75.18\pm0.52$  & & $64.32\pm0.81$ & $65.28\pm0.52$ &  & $71.03\pm0.73$  & $74.08\pm0.65$  & & $86.12\pm0.37$ & $90.29\pm0.38$  \\
\midrule

Meta-PN  & $\textbf{74.94}\pm\textbf{0.25}$  & $\textbf{79.88}\pm\textbf{0.15}$  & & $\textbf{70.48}\pm\textbf{0.34}$ & $\textbf{74.14}\pm\textbf{0.50}$ &  & $\textbf{73.25}\pm\textbf{0.77}$ & $\textbf{77.78}\pm\textbf{0.92}$  &  & $\textbf{88.99}\pm\textbf{0.29}$ & $\textbf{91.31}\pm\textbf{0.22}$   \\

\bottomrule
\end{tabular}}

\label{table:semi}

\end{table*}

\smallskip
\noindent \textbf{Compared Methods.} To corroborate the effectiveness of our approach, three categories of baselines are included in our experiments: (i) \textit{Classical Models}. \textbf{MLP}, \textbf{LP} (Label Propagation)~\cite{zhou2004learning} are two classical models using only feature and structure information, respectively. \textbf{GCN}~\cite{kipf2017semi} and \textbf{SGC}~\cite{wu2019simplifying} are two representative GNN models. Due to the space limit, we omit some baselines like GAT, GraphSAGE since similar results can be observed; (ii) \textit{Label-efficient GNNs}. \textbf{GLP} (Generalized Label Propagation) and \textbf{IGCN} (Improved GCN)~\cite{li2019label} are two models combine label propagation and GCN from a unifying graph filtering perspective. \textbf{M3S} \cite{sun2020multi} is a multi-stage self-training framework, which incorporates self-supervised learning to improve the model performance with few labeled nodes; (iii) \textit{Deep GNNs}. \textbf{APPNP}~\cite{klicpera2019predict} decouples prediction and propagation with performing personalized propagation of neural predictions, while  \textbf{DAGNN}~\cite{liu2020towards} adaptively incorporate information from large receptive fields. \textbf{C\&S}~\cite{huang2021combining} is an effective model that combines label propagation and simple neural networks.
 \textbf{GPR-GNN}~\cite{chien2021adaptive} addresses the limitation of APPNP on different types of graphs with adaptive propagation weights. 

\smallskip
\noindent{\textbf{Implementation Details.}} 
All our experiments are conducted with a 12 GB Ti-tan Xp GPU. The proposed Meta-PN is implemented in PyTorch. We use a 2-layer MLP with 64 hidden units for the feature-label transformer. We apply L2 regularization with $\lambda = 0.005$ on the weights of the first neural layer and set the dropout rate for both neural layers to be 0.3. For methods based on label propagation, we use $K = 10$ power iteration (propagation) steps by default. To make a fair comparison, we let all the configurations of the baselines be the same as Meta-PN including neural network layers, hidden units, regularization, propagation steps, early stopping and initialization. We use Adam to optimize the baseline methods as suggested and fine-tune for the corresponding learning rate on different datasets. More details on model implementation and parameter selection can be found in Appendix \ref{detail}.

\begin{figure*}[!t]
    \graphicspath{{figures/}}
    \centering
    \subfigure[label ratio $= 1.0\%$] 
    {
    \includegraphics[width=0.22\textwidth]{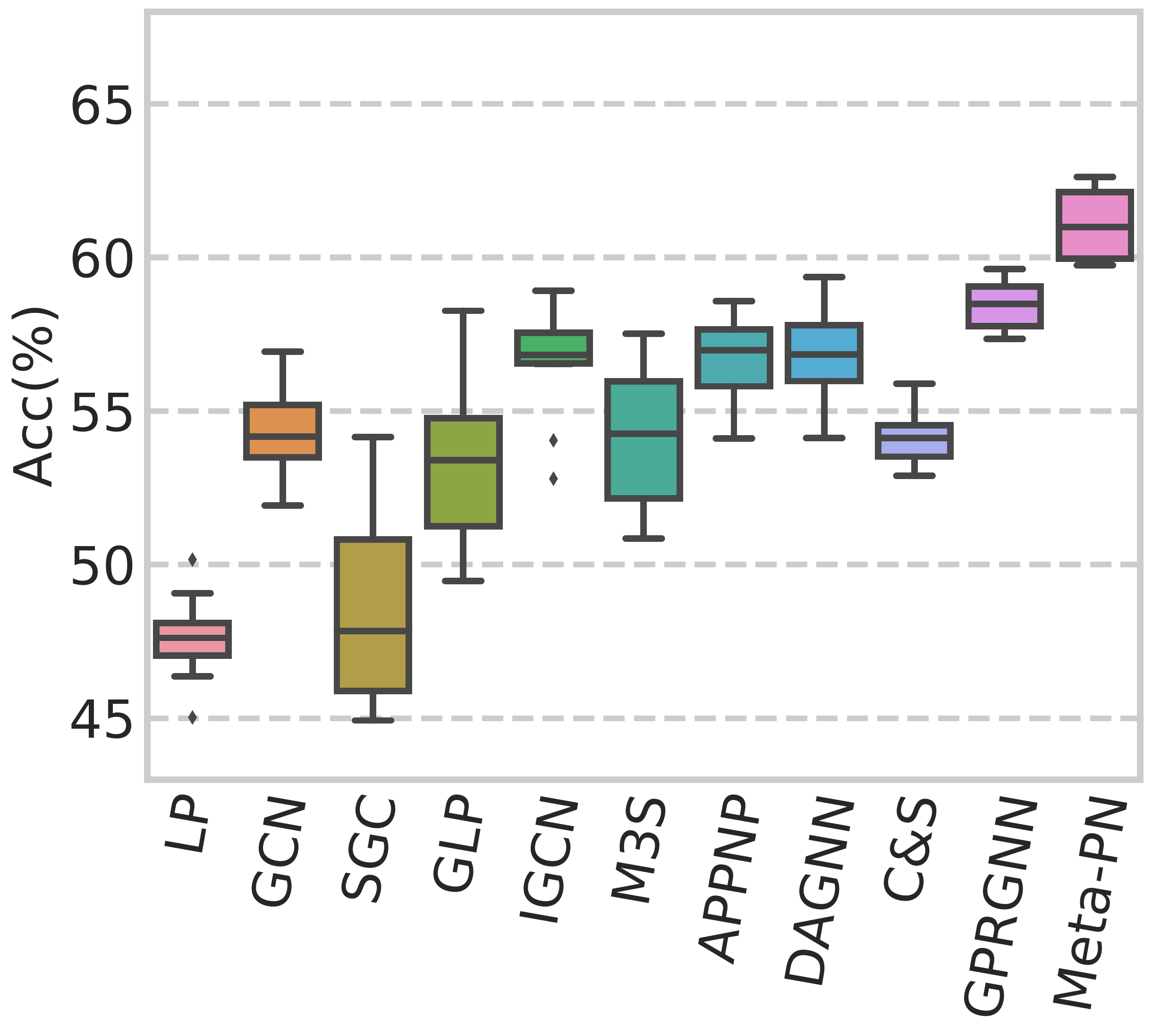}
    }
   \hspace{-0.25cm}
    \subfigure[label ratio $= 1.5\%$]
    {
    \includegraphics[width=0.22\textwidth]{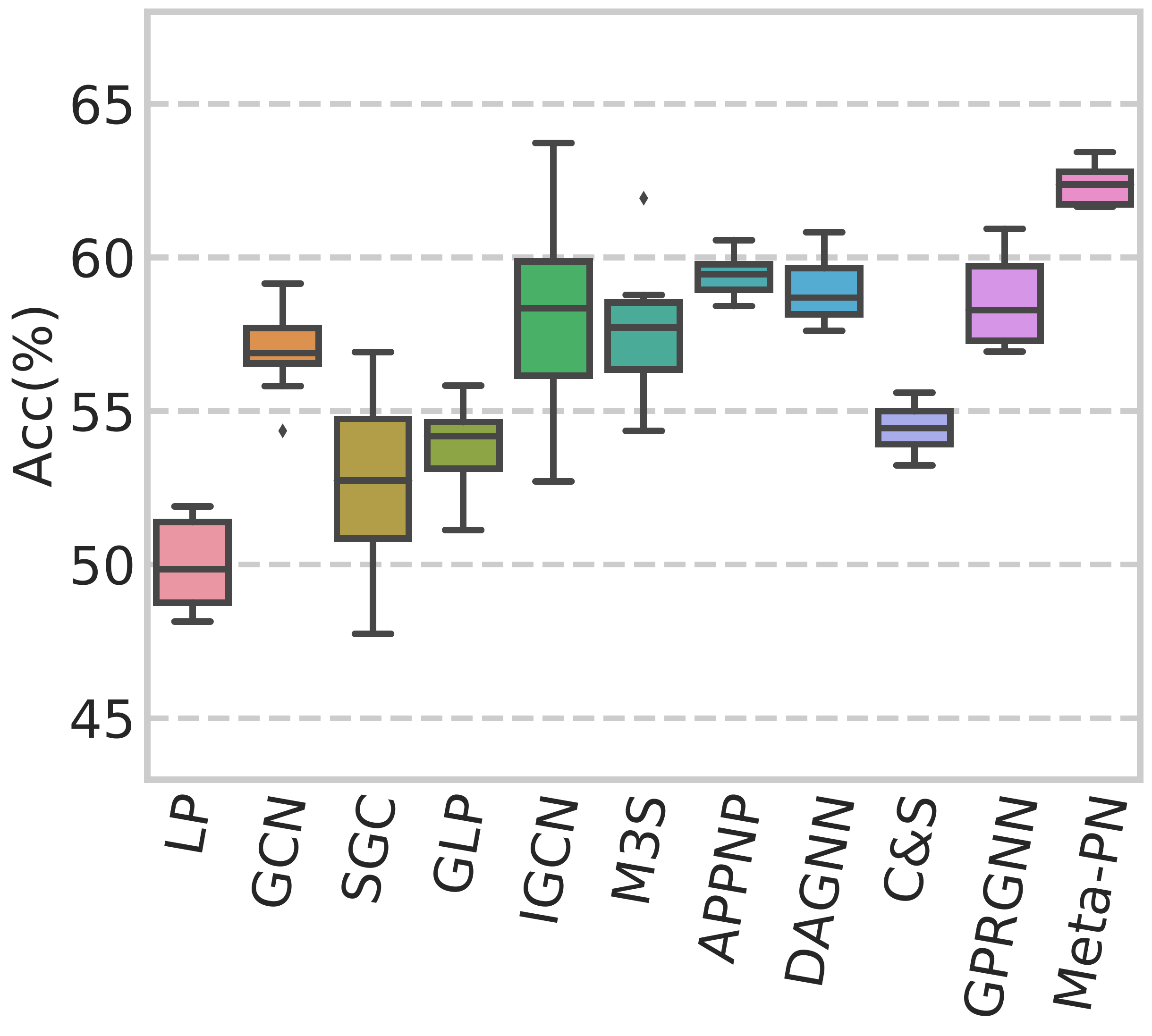}
    }
    \hspace{-0.25cm}
    \subfigure[label ratio $= 2.0\%$] 
    {
    \includegraphics[width=0.22\textwidth]{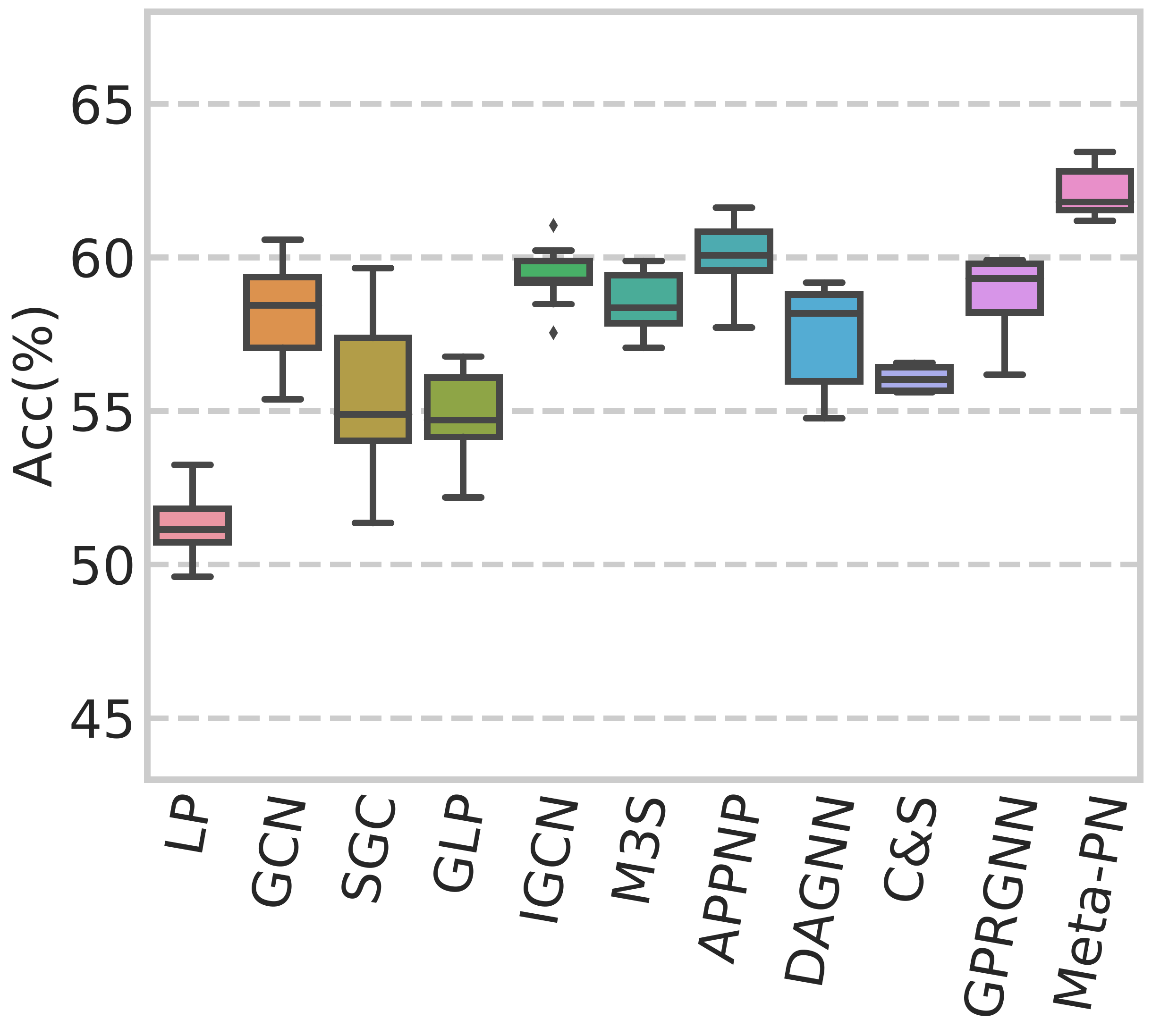}
    }
    \hspace{-0.25cm}
    \subfigure[label ratio $= 2.5\%$]
    {
    \includegraphics[width=0.22\textwidth]{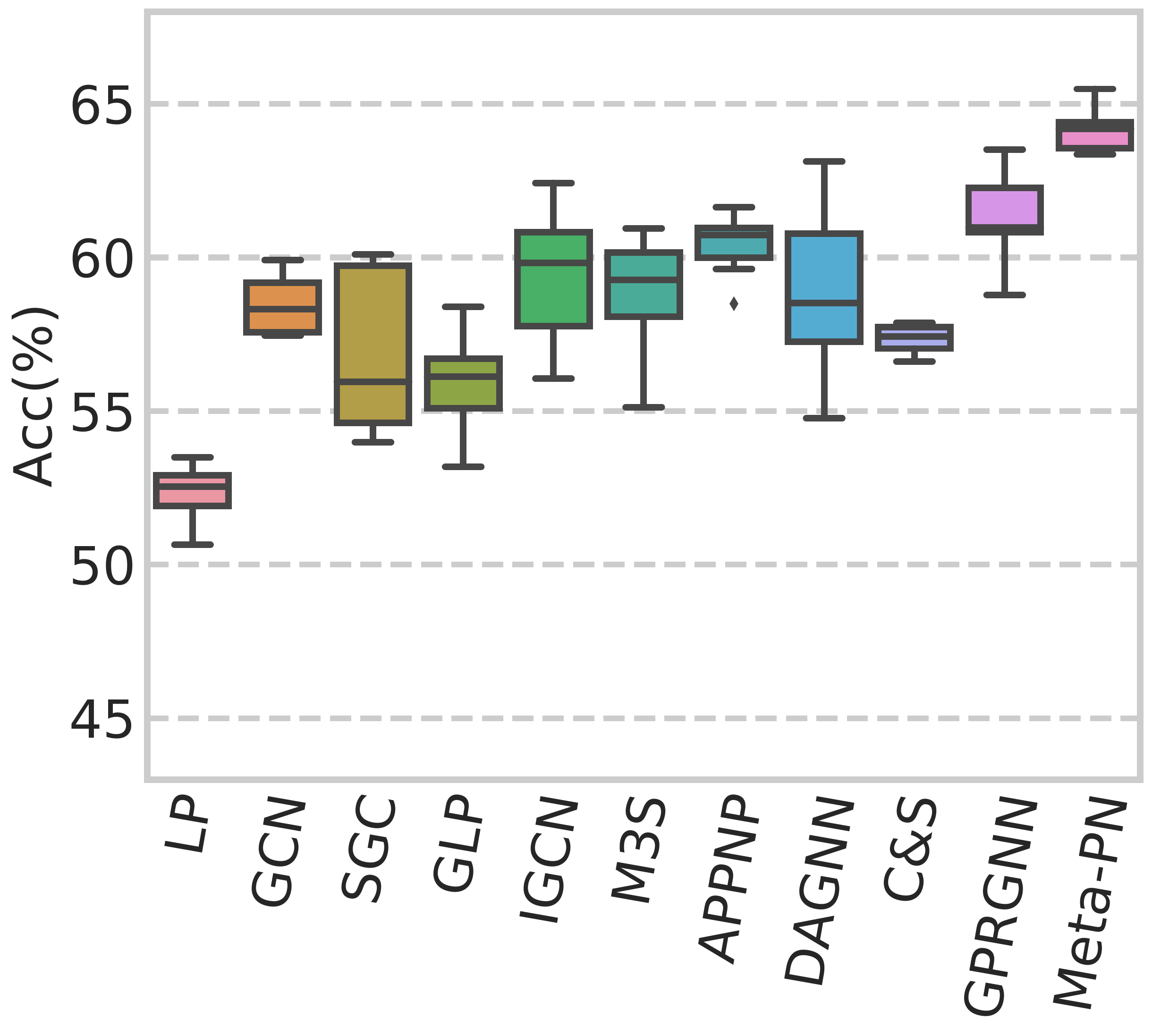}
    }
    \caption{Comparison results on ogbn-arxiv w.r.t different size of training labels.}%

    \label{fig:arxiv_results}
\end{figure*}

\subsection{Evaluation Results}

\begin{figure*}[t!]
    \graphicspath{{figures/}}
    \centering
    \subfigure[Cora-ML] 
    {
    \includegraphics[width=0.22\textwidth]{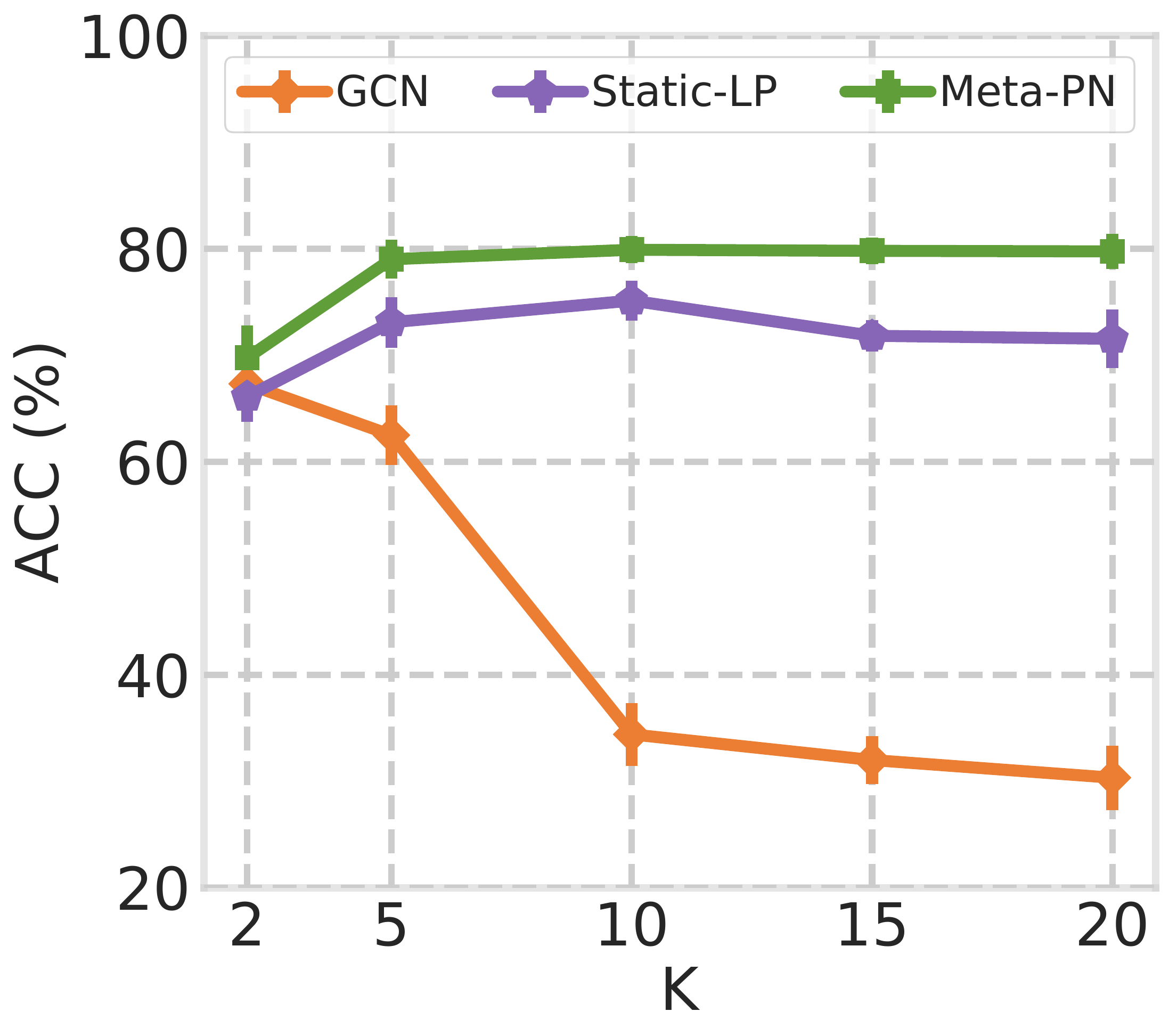}
    }
   \hspace{-0.2cm}
    \subfigure[CiteSeer]
    {
    \includegraphics[width=0.22\textwidth]{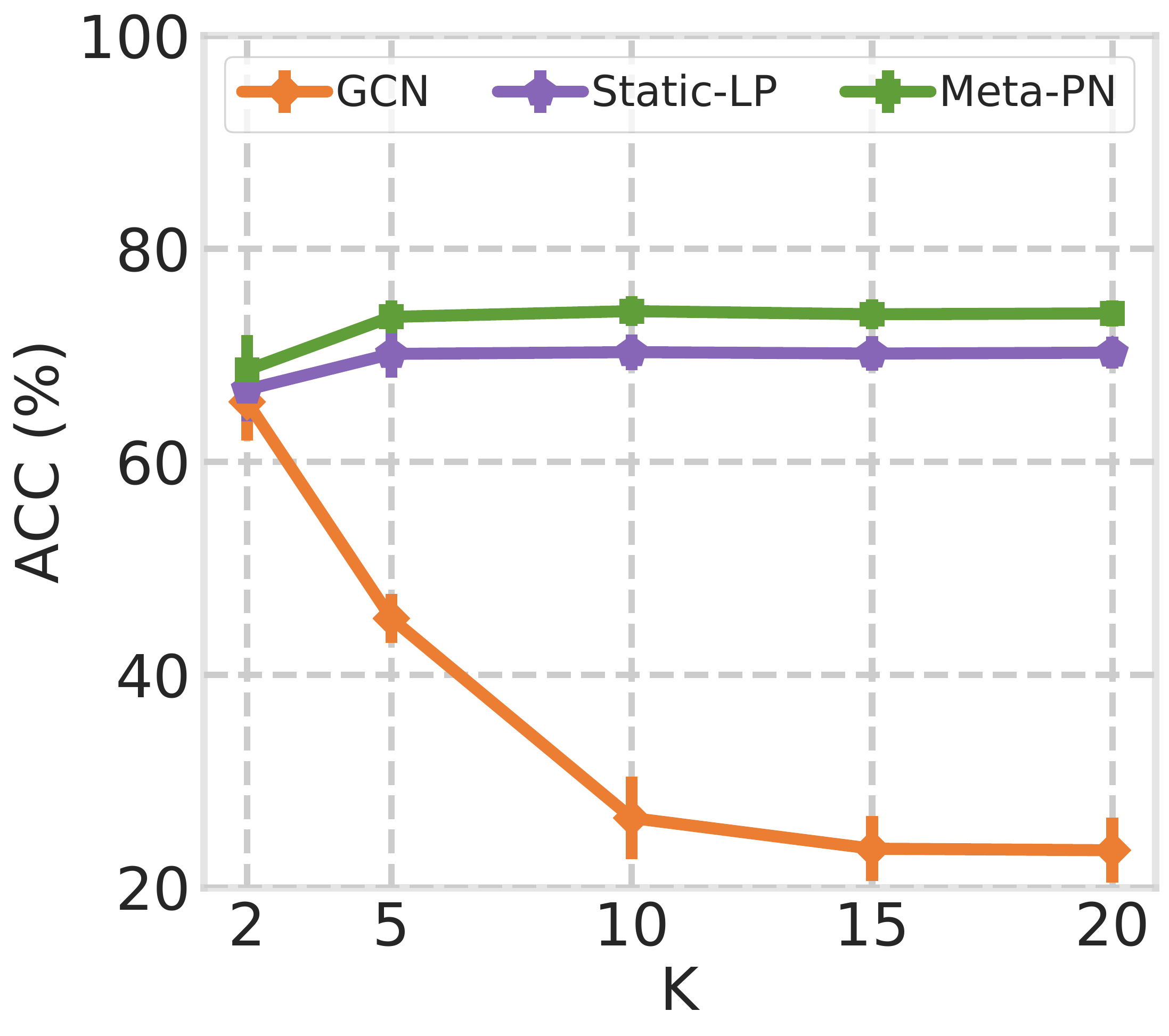}
    }
    \hspace{-0.2cm}
    \subfigure[PubMed] 
    {
    \includegraphics[width=0.22\textwidth]{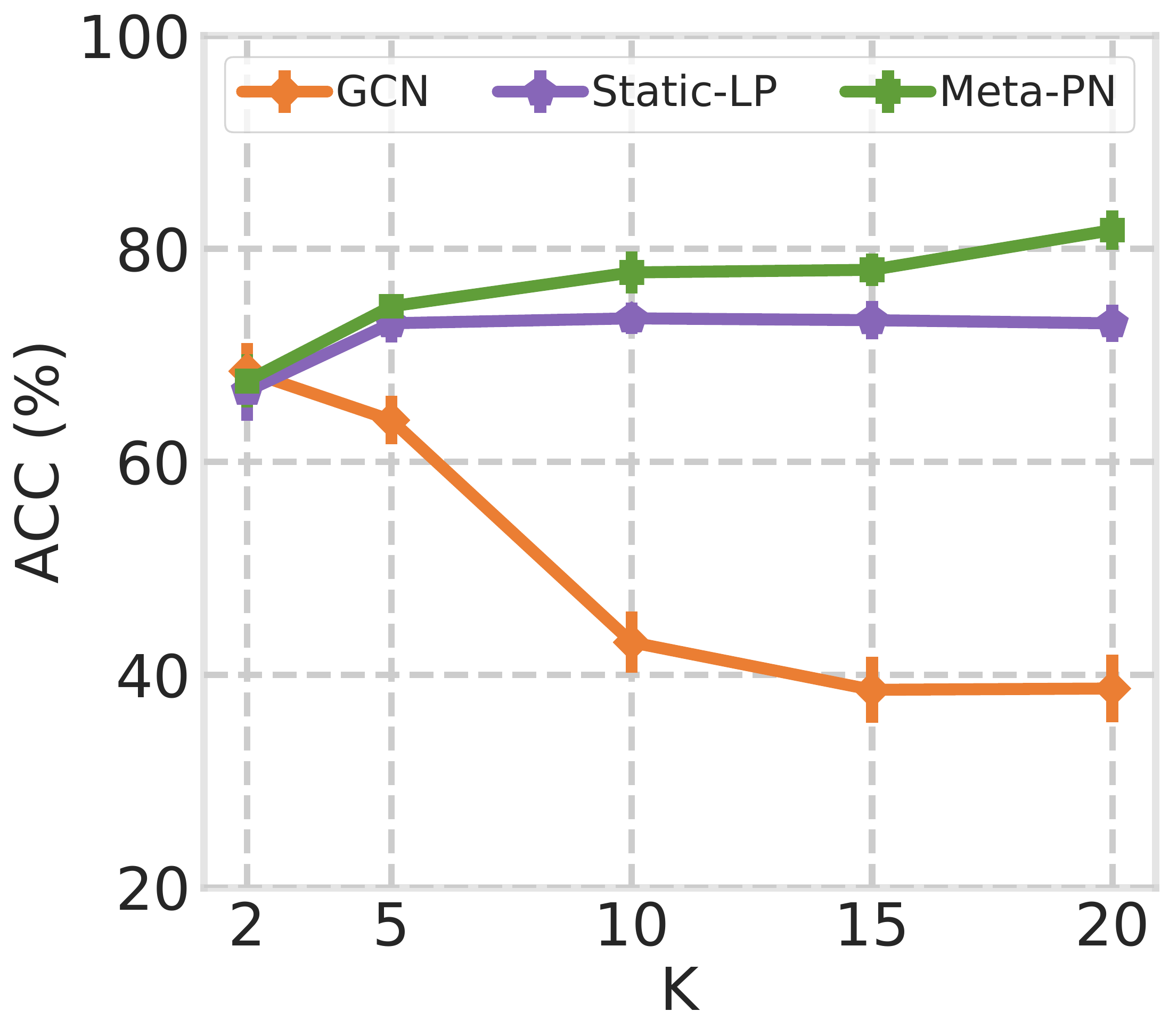}
    }
    \hspace{-0.2cm}
    \subfigure[Ogbn-arxiv]
    {
    \includegraphics[width=0.22\textwidth]{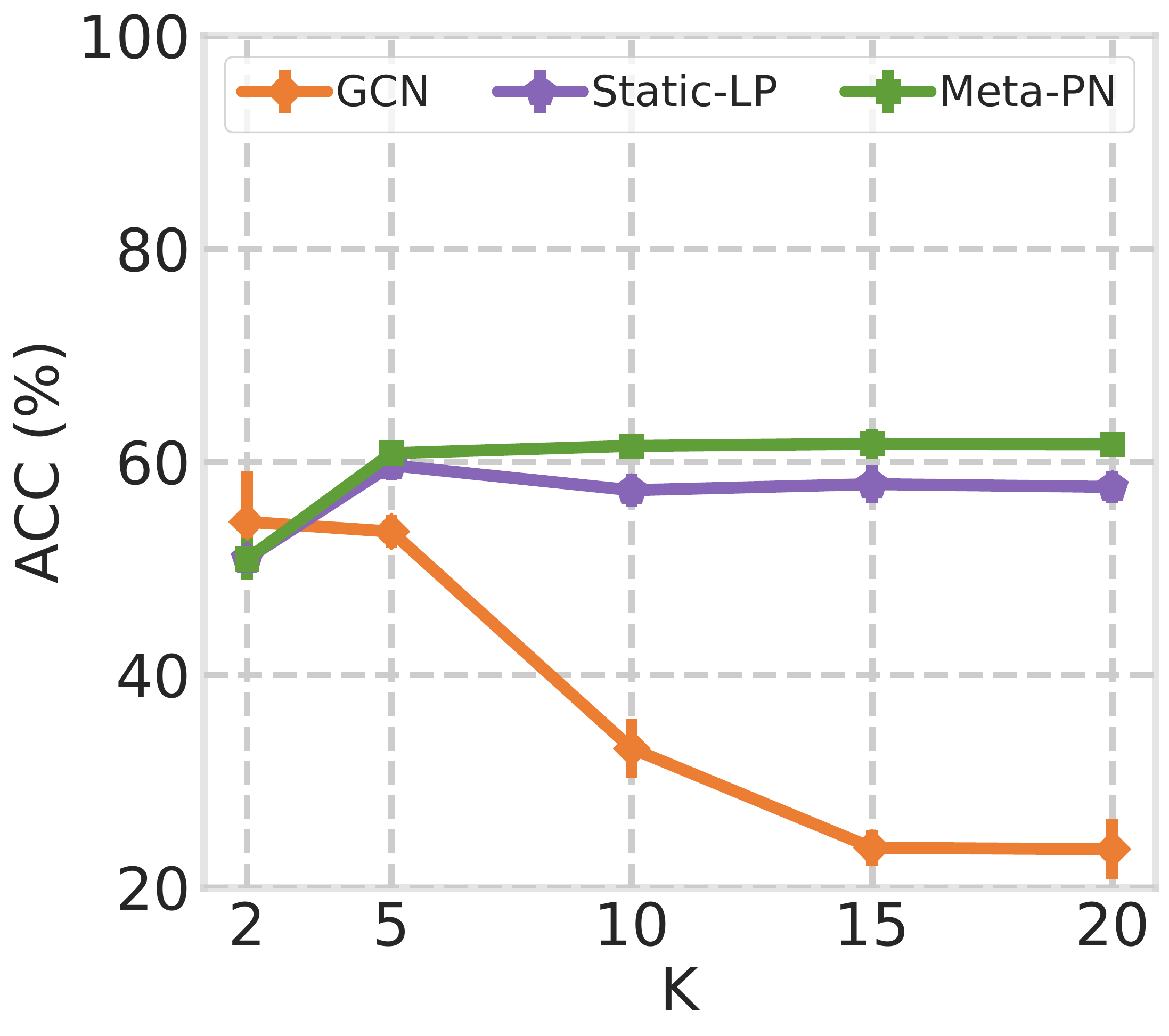}
    }
    \caption{Few-shot (i.e., 5-shot or 1.0\% label ratio) evaluation on different datasets w.r.t. propagation steps (K).}%

    \label{fig:kstep}
\end{figure*}

\begin{table}[b!]
\centering
\caption{Test accuracy on standard semi-supervised node classification: mean accuracy ($\%$) $\pm$ 95$\%$ confidence interval.}
\scalebox{0.82}{
\begin{tabular}{@{}lccccccccc@{}}
\toprule

\rule{0pt}{10pt} \textbf{Method} & \multicolumn{1}{c}{Cora-ML}  & \multicolumn{1}{c}{CiteSeer}  &  \multicolumn{1}{c}{PubMed} &  \multicolumn{1}{c}{MS-CS}

\\ \midrule

MLP       & $68.42\pm.34$   & $63.98\pm .44$   &  $69.47\pm.47$ & $88.30\pm.13$  \\
LP        & $75.74\pm.27$  & $65.62\pm.43$  &  $69.82\pm.70$ & $72.03\pm.25$  \\

GCN    & $82.70\pm.37$  & $73.62\pm.39$  &  $76.84\pm.44$ & $91.10\pm.20$  \\

SGC    & $75.97\pm.72$  & $75.57\pm.28$  &  $71.24\pm.86$ & $90.56\pm.14$    \\
\midrule

GLP   & $ 81.67\pm.14$  & $75.21\pm.14$  &  $78.95\pm.09$ & $91.85\pm.04$   \\
IGCN   & $ 82.11\pm.09$  & $75.22\pm.10$  &  $79.06\pm.07$ & $91.60\pm.03$   \\
M3S   & $82.72\pm.13$  & $73.73\pm.32$  &  $77.62\pm.11$ & $91.08\pm.09$   \\
\midrule
APPNP   & $85.09\pm.25$  & \underline{$75.73\pm.30$}  &  \underline{$79.73\pm.31$} & $91.74\pm.16$  \\
DAGNN   & \underline{$85.65\pm.23$}  & $74.53\pm.17$  &  $79.59\pm.37$ & \underline{$92.80\pm.17$} \\
C\&S  & $83.18\pm.31$  & $70.51\pm.24$  &  $77.10\pm.34$ & $92.49\pm.19$ \\
GPR-GNN   & $83.53\pm.31$  & $71.18\pm.25$  &  $79.62\pm.46$ & $92.57\pm.21$ \\
\midrule

Meta-PN  & $\textbf{86.33}\pm\textbf{.36}$  & $\textbf{77.13}\pm\textbf{.31}$  &  $\textbf{80.39}\pm\textbf{.53}$ & $\textbf{93.92}\pm\textbf{.17}$ \\

\bottomrule
\end{tabular}}
\label{table:20-shot}

\end{table}

\noindent{\textbf{Few-shot Semi-supervised Evaluation.}} First, we evaluate the proposed approach Meta-PN and all the baseline methods on few-shot semi-supervised node classification, which aims to predict the missing node labels with only a few labeled nodes. The average test accuracies under the few-shot setting (i.e., 3-shot and 5-shot) can be found in Table~\ref{table:semi}. Additional results are provided in Appendix \ref{addition} due to the space limit. From the reported results, we can clearly see that Meta-PN significantly outperforms all the baseline methods on each dataset based on paired t-tests with $p < 0.05$. Specifically, we elaborate our in-depth observations and analysis as follows: 
(i) without abundant labeled data, classical models including vanilla GNNs only obtain very poor classification accuracy under different evaluation entries; 
(ii) overall the label-efficient GNNs outperform classical GNNs, but still cannot achieve satisfying results. One major reason is that those methods cannot handle the oversmoothing issue since they are incapable of explicitly leveraging the knowledge from large receptive fields; (iii) by enabling better propagation of label signals, deep GNNs have stronger performance than both the classical models and label-efficient GNNs, which again demonstrates the necessity of addressing the oversmoothing issue for solving the few-shot semi-supervised learning problem. However, existing deep GNNs are not specifically developed to tackle the data sparsity issue, thus their performance still falls behind Meta-PN by a noticeable margin on different datasets when only very few labels are available. This observation proves that Meta-PN is able to address the overfitting and oversmoothing issues when labeled data is extremely sparse by combining the power of large receptive fields and pseudo labels.



    
\smallskip
\noindent{\textbf{Standard Semi-supervised Evaluation.}} To make our evaluation more comprehensive, we then examine the effectiveness of Meta-PN under the standard semi-supervised node classification tasks. As in \cite{klicpera2019predict}, we randomly sample 20 labeled nodes for each class (i.e., 20-shot) as the training set. According to the average performance reported in Table~\ref{table:20-shot}, we make the following observations: 
(i) the GNN models which combine both the structure and feature knowledge from labeled nodes can obtain improved node classification performance compared to methods which only consider feature or structure information individually; (ii) under the standard semi-supervised node classification task, the performance of the label-efficient GNNs are close to vanilla GNNs; (iii) though Meta-PN is mainly proposed for few-shot semi-supervised learning, it still achieves the best performance for the standard semi-supervised node classification task, illustrating the superiority of our graph approach.


\smallskip
\noindent\textbf{Evaluation on Open Graph Benchmark (OGB).} Real-world graphs commonly have a larger size and more node classes than many toy graphs, leading to the collected graphs having noisy structures and complex properties. To further illustrate the effectiveness of our approach on large-scale real-world graphs, we adopt the widely used ogbn-arxiv dataset and compare all the methods under the few-shot setting  (i.e., from 1\% to 2.5\% label ratio). We summarize their performance for few-shot semi-supervised node classification on ogbn-arxiv in Figure \ref{fig:arxiv_results} by changing the ratio of training labels, in which we omit MLP as its test accuracy is much lower than the other methods. We can observe that Meta-PN can significantly outperform all the baseline models under different few-shot environments. Compared to the other baseline methods, the performance of Meta-PN is relatively stable when we decrease the ratio of training labels, which demonstrates the robustness of Meta-PN in handling noisy and complex real-world graphs. Remarkably, our approach can achieve close performance to the vanilla GCN on ogbn-arxiv with much fewer labeled nodes (2.5\% vs. 54\%).

\smallskip
\noindent{\textbf{Parameter \& Ablation Analysis.}} 
To demonstrate the effects of using different propagation steps and the importance of the meta-leaned label propagation strategy for Meta-PN, we compare our approach with two baselines under the $5$-shot (or 1.0\% label ratio for ogbn-arxiv) semi-supervised setting with varying number of propagation steps. Specifically, \textit{GCN} learns the node representation with the standard message-passing scheme while \textit{Static-LP} representing the variant of Meta-PN that uses fixed teleport probabilities instead of meta-learned ones. The evaluation results are shown in Figure \ref{fig:kstep}. As we can observe from the figure, \textit{GCN} can achieve very close performance with the other two methods when the number of propagation steps is relatively small. While if we largely increase the number of propagation steps, the performance of \textit{GCN} breaks down due to the oversmoothing issue. Empowered by the idea of label propagation, \textit{Static-LP} can largely alleviate the oversmoothing issue and significantly outperform \textit{GCN}. This verifies that larger propagation steps or receptive fields are necessary for improving the performance of GNN when labeled data is extremely limited. In the meantime, \textit{Static-LP} still falls behind Meta-PN, mainly because of the infeasibility of balancing the importance of different receptive fields. On the contrary, Meta-PN is able to address this issue by inferring optimal pseudo labels on unlabeled nodes with our meta-learning algorithm. Its performance becomes stable when $K \ge 10$, indicating that Meta-PN can obtain good performance considering both efficiency and effectiveness with a moderate number of propagation steps (e.g., $K = 10$). 

\section {Conclusion}
\label{sec:conclusion}

In this paper, we propose a new graph meta-learning framework, Meta Propagation Networks (Meta-PN), for solving the problem of few-shot semi-supervised node classification. Based on the meta-learned label propagation strategy, we are able to generate informative pseudo labels on unlabeled nodes, in order to augment the insufficient labeled data and learn a powerful GNN model. Though built with simple neural networks, Meta-PN effectively enables larger receptive fields and avoids oversmoothing when learning with very few labeled data. We test Meta-PN on a spectrum of benchmark datasets and the results well demonstrate its effectiveness. For future work, it would be interesting to investigate other pseudo-labeling strategies for solving the studied problem.

\clearpage
\section*{Acknowledgements}
This work is partially supported by Office of Naval Research (ONR) N00014-21-1-4002 and Army Research Office (ARO) W911NF2110030.
\bibliography{aaai22}

\newpage
\appendix
\section{Appendix}
\subsection{Theoretical Analysis of Meta-PN}
\label{proof}
As we introduced in Section 3.1, the target model trained with meaningful pseudo labels  can be considered as a special variant of GCN, which allows far more propagation steps with much fewer parameters. Here we present the theoretical analysis to show this connection. For simplicity, we first consider the 1-layer (one propagation step) case, and we have the cross-entropy loss function as follows:
\begin{equation}
\begin{aligned}
     \mathcal{L}_1 &=  \mathcal{L}_{\text{CE}}(\mathbf{T} f_{\bm\theta_1}(\mathbf{X}), \mathbf{Y})\\
     &= - \sum_{j \in \mathcal{V}^L, k \in \mathcal{C}} y_{j,k}(\log \sum_{i \in \mathcal{V}} t_{j,i} p_{i,k}),
\end{aligned}
\end{equation}
where $\mathbf{p}_{i}$ represents the result from the transformation function $f_{\bm\theta_1}(\mathbf{x}_i)$ of GCN and $p_{i,k}$ denotes its $k$-th element. $\mathcal{L}_{\text{CE}}$ denotes the cross-entropy loss. The gradients of the objective function with respect to $\bm\theta_1$ can be written as:
\begin{equation}
\begin{aligned}
    \nabla_{\bm\theta_1} \mathcal{L}_1 &= - \sum_{j \in \mathcal{V}^L, k \in \mathcal{C}} y_{j,k} \nabla_{\bm\theta_1} (\log \sum_{i \in \mathcal{V}} t_{j,i} p_{i,k})\\
    &= - \sum_{j \in \mathcal{V}^L, k \in \mathcal{C}} y_{j,k}  \frac{\sum_{i \in \mathcal{V}} t_{j,i} \nabla_{\bm\theta_1} p_{i,k}}{\sum_{q \in \mathcal{V}} t_{j,q} p_{q,k}}.
\end{aligned}
\end{equation}

Note that $\mathbf{y}_j$ is an one-hot vector, only the $h(j)$-th element is non-zero. Then the gradients can be rewritten as follows:
\begin{equation}
\begin{aligned}
    \nabla_{\bm\theta_1} \mathcal{L}_1 &= - \sum_{j \in \mathcal{V}^L} y_{j,h(j)}  \frac{\sum_{i \in \mathcal{V}} t_{j,i} \nabla_{\bm\theta_1} p_{i,h(j)}}{\sum_{q \in \mathcal{V}} t_{j,q} p_{q,h(j)}}\\
    &= - \sum_{i\in \mathcal{V}, j \in \mathcal{V}^L} \frac{ t_{j,i} p_{i,h(j)}}{\sum_{q \in \mathcal{V}} t_{j,q} p_{q,h(j)}}  y_{j,h(j)}  \frac{\nabla_{\bm\theta_1} p_{i, h(j)}}{p_{i, h(j)}}\\
    &= \sum_{i\in \mathcal{V}, j \in \mathcal{V}^L} \frac{ t_{j,i} p_{i,h(j)}}{\sum_{q \in \mathcal{V}} t_{j,q} p_{q,h(j)}} \nabla_{\bm\theta_1} \mathcal{L}_{\text{CE}}(\mathbf{p}_i, \mathbf{y}_j).
    \label{eq:decoupled}
\end{aligned}
\end{equation}


If we train a neural network $f_{\bm\theta_2}(\cdot)$ on the directly propagated pseudo labels $\hat{\mathbf{Y}} = \mathbf{T}\mathbf{Y}$, the gradients of its objective function can be computed as follows:
\begin{equation}
\begin{aligned}
   \nabla_{\bm\theta_2} \mathcal{L}_2 &=  \nabla_{\bm\theta_2} \mathcal{L}_{\text{CE}}( f_{\bm\theta_2}(\mathbf{X}), \mathbf{T}\mathbf{Y})\\
   &= \sum_{i\in \mathcal{V}, k \in \mathcal{C}} \sum_{j \in \mathcal{V}^L} t_{i,j} y_{j, k} \nabla_{\bm\theta_2} \log  p_{i, k}\\
   &= \sum_{i\in \mathcal{V}, j \in \mathcal{V}^L} t_{i, j} \nabla_{\bm\theta_2} \sum_{k \in \mathcal{C}} y_{j, k} \log p_{i, k}\\
   &= \sum_{i\in \mathcal{V}, j \in \mathcal{V}^L}  t_{i, j} \nabla_{\bm\theta_2} \mathcal{L}_{\text{CE}}(\mathbf{p}_i, \mathbf{y}_j).
\end{aligned}
\end{equation}
As we can see, the learning process of $f_{\bm\theta_2}(\cdot)$ is equivalent to  $f_{\bm\theta_1}(\cdot)$ by ignoring the regularization term $\frac{ p_{i,h(j)}}{\sum_{q \in \mathcal{V}} t_{j,q} p_{q,h(j)}}$ in Eq. (\ref{eq:decoupled}). For Meta-PN, this regularization term is replaced by the teleport probability $\alpha$, thus the learned target model is essentially a special case of GCN.

\subsection{Computing \texorpdfstring{$\nabla_{\bm\phi} J_{\text{gold}}(\bm\theta'(\bm\phi))$}.}
\label{gradient}

Due to the relationship between $\bm\theta$ and $\bm\phi$ as shown in Eq. (\ref{eq:bi-level}), $\nabla_{\bm\phi} J_{\text{gold}}(\bm\theta'(\bm\phi))$ is differentiable with respect to $\bm\phi$. Then we first compute the gradient $\nabla_{\bm\phi} J_{\text{gold}}(\bm\theta'(\bm\phi))$ by applying chain rule with implicit and explicit gradients as follows:
\begin{equation}
\begin{aligned}
     \nabla_{\bm\phi} J_{\text{gold}}(\bm\theta'(\bm\phi)) &=
    \nabla_{\bm\phi} J_{\text{gold}} (\bm\theta - \eta_{\bm\theta} \nabla_{\bm\theta} J_{\text{pseudo}}(\bm\theta, \bm\phi))\\
     &=  - \eta_{\bm\theta} \nabla_{\bm\theta, \bm\phi}^2 J_{\text{pseudo}}(\bm\theta, \bm\phi) \nabla_{\bm\theta'} J_{\text{gold}}(\bm\theta'(\bm\phi)),
     \label{eq:chainrule}
\end{aligned}
\end{equation}
where $\nabla_{\bm\phi} J^+_{\text{gold}}(\bm\theta'(\bm\phi))$ is the implicit gradient that assumes all other variables except $\bm\phi$ as constants.


The second term of the above equation contains an expensive matrix-vector product, here we use the finite difference approximation to reduce the complexity. Let $\epsilon$ denote a small constant number, according to the finite difference method, $\frac{\partial f(x,z)}{\partial x} * k \approx \lim_{\epsilon \to 0} \frac{f(x + \epsilon, z) - f(x - \epsilon, z)}{2\epsilon} * k = \lim_{\epsilon \to 0} \frac{f(x + k\epsilon, z) - f(x - k\epsilon, z)}{2k\epsilon} * k = \lim_{\epsilon \to 0} \frac{f(x + k\epsilon, z) - f(x - k\epsilon, z)}{2\epsilon}$. Thus we can get:
\begin{equation}
\begin{aligned}
    &\nabla_{\bm\theta, \bm\phi}^2 J_{\text{pseudo}}(\bm\theta, \bm\phi) \nabla_{\bm\theta'} J_{\text{gold}}(\bm\theta'(\bm\phi))\\
    &= \nabla_{\bm\phi}\Big(\nabla_{\bm\theta} J_{\text{pseudo}}(\bm\theta, \bm\phi) * \nabla_{\bm\theta'} J_{\text{gold}}(\bm\theta'(\bm\phi))\Big)\\
    &\approx 
    \lim_{\epsilon \to 0} \frac{\nabla_{\bm\phi}J_{\text{pseudo}} \Big(\bm\theta + \nabla_{\bm\theta'} J_{\text{gold}}(\bm\theta'(\bm\phi)) \epsilon, \bm\phi\Big)}{2\epsilon}\\
    &- \frac{\nabla_{\bm\phi}J_{\text{pseudo}} \Big(\bm\theta - \nabla_{\bm\theta'} J_{\text{gold}}(\bm\theta'(\bm\phi)) \epsilon, \bm\phi\Big)}{2\epsilon}.
    \label{eq:finite}
\end{aligned}
\end{equation}

With Eq. (\ref{eq:chainrule}) and Eq. (\ref{eq:finite}), we can get:
\begin{equation}
\begin{aligned}
    \nabla_{\bm\phi} J_{\text{gold}}(\bm\theta'(\bm\phi))
    &= -  
    \eta_{\bm\phi} \lim_{\epsilon \to 0} \frac{\nabla_{\bm\phi} J_{\text{pseudo}}(\bm\theta^+, \bm\phi) - \nabla_{\bm\phi} J_{\text{pseudo}}(\bm\theta^-, \bm\phi)}{2\epsilon}\\
    &=  - \frac{\eta_{\bm\phi}}{2\epsilon}[\nabla_{\bm\phi} J_{\text{pseudo}}(\bm\theta^+, \bm\phi) - \nabla_{\bm\phi} J_{\text{pseudo}}(\bm\theta^-, \bm\phi)],
\end{aligned}
\end{equation}
where $\bm\theta^{\pm} = \bm\theta \pm \nabla_{\bm\theta'} J_{\text{gold}}(\bm\theta'(\bm\phi)) \epsilon$.

\subsection{Implementation Details}
\label{detail}

\noindent\textbf{Meta-PN.} We implement the proposed Meta-PN in PyTorch. We set the batch size to 1,024 for Cora-ML and Citeseer, and 4,096 for the other datasets. Specifically, we use two-layer MLP with 64 hidden units for the \textit{feature-label transformer} and optimize it with Adam. We grid search for the learning rate $\eta_{\bm\theta}$ in \{$1 \times 10^{-5}$, $5 \times 10^{-5}$, $1 \times 10^{-4}$, $5 \times 10^{-4}$, $1 \times 10^{-3}$, $5 \times 10^{-3}$, $1 \times 10^{-2}$, $5 \times 10^{-2}$, $1 \times 10^{-1}$, $5 \times 10^{-1}$\}. 
Meanwhile, we optimize the \textit{adaptive label propagator} with Adam and grid search for the learning rate $\eta_{\bm\phi}$ in \{$1 \times 10^{-5}$, $5 \times 10^{-5}$, $1 \times 10^{-4}$, $5 \times 10^{-4}$, $1 \times 10^{-3}$, $5 \times 10^{-3}$, $1 \times 10^{-2}$, $5 \times 10^{-2}$, $1 \times 10^{-1}$, $5 \times 10^{-1}$\}.
We also search for dropout rate in $\{0.1, 0.2, 0.3, 0.4, 0.5, 0.6, 0.7\}$. The optimal values are selected when the model achieve the best performance for validation set. Following~\cite{klicpera2019predict}, the early stopping criterion uses a patience of $p = 100$ and an (unreachably high) maximum of $n = 10,000$ epochs. The patience is reset whenever the accuracy increases or the loss decreases on the validation set.

\smallskip
\noindent\textbf{Baselines.} In our experiments, we compare our approach with different methods including MLP, LP, GCN, SGC, GLP, IGCN, M3S, APPNP, DAGNN, C\&S and GPR-GNN. For the baseline methods, we adopt their public implementations and the details are as follows:
\begin{itemize}
    \item \textbf{MLP}: For a fair comparison, we use a 2-layer fully connected network with 64 hidden units for representation learning.
    \item \textbf{LP}~\cite{zhou2004learning}: We use the same propagation step K and the teleport probability as Meta-PN for a fair comparison.
    \item \textbf{GCN}\footnote{ \url{https://github.com/tkipf/pygcn}} \cite{kipf2017semi}: We build the GCN model with two graph convolutional layers (64 dimensions) for learning node representations.
    \item \textbf{SGC}\footnote{\url{https://github.com/Tiiiger/SGC}} \cite{wu2019simplifying}: After the feature pre-processing step, it learns the node representations with 2-layer feature propagation with 64 hidden units. 
    \item \textbf{GLP \& IGCN}\footnote{\url{https://github.com/liqimai/Efficient-SSL}} \cite{li2019label}:  It uses a two-layer structure (64 hidden units) in which the filter parameters $k$ and $\alpha$ is set to be 5 and 10 for 20-shot, and is set to be 10 and 20 for all the other tasks. The results with the best performing filter (i.e., RNM or AR) are reported.
    \item \textbf{M3S}\footnote{\url{https://github.com/datake/M3S}} \cite{sun2020multi}:  We fix the number of clusters as 200 and select the best number of layers and stages as suggested by the authors. 
   \item \textbf{APPNP}\footnote{\url{https://github.com/klicperajo/ppnp}}~\cite{klicpera2019predict}: Similar to Meta-PN, we use the 2-layer MLP (64 hidden units), with 10 steps of propagation. For the best performance, we set the teleport probability $\alpha = 0.1$ for the citation graphs and use $\alpha = 0.2$ for the co-authorship graph due to their structural difference. 
    \item \textbf{DAGNN}\footnote{\url{https://github.com/mengliu1998/DeeperGNN}} \cite{liu2020towards}: We let the size of hidden unit and the propagation step to be the same as Meta-PN for fairness.
    \item \textbf{C\&S}\footnote{\url{https://github.com/CUAI/CorrectAndSmooth}} \cite{huang2021combining}: We use the MLP base predictor and follow the default settings provided by the authors for the best performance.
    \item \textbf{GPR-GNN}\footnote{\url{https://github.com/jianhao2016/GPRGNN}} \cite{chien2021adaptive}: For fair comparison, we use the random walk path lengths with K = 10 and use a 2-layer (MLP) with 64 hidden units for the neural network component.
    
\end{itemize}

For all the baseline methods, we use Adam as optimizer and fine-tune the hyperparameters on each dataset. Specifically, we grid search for the learning rate in \{$1 \times 10^{-5}$, $5 \times 10^{-5}$, $1 \times 10^{-4}$, $5 \times 10^{-4}$, $1 \times 10^{-3}$, $5 \times 10^{-3}$, $1 \times 10^{-2}$, $5 \times 10^{-2}$, $1 \times 10^{-1}$, $5 \times 10^{-1}$ \} and dropout rate in $\{0.1, 0.2, 0.3, 0.4, 0.5, 0.6, 0.7\}$. Also, we use the same early stop strategy as for Meta-PN. 

\smallskip
\noindent\textbf{Packages Used for Implementation.} For reproducibility, we also list the packages we use in the implementation with their corresponding versions: python==3.6.6, pytorch==1.4.0, cuda==10.1, numpy==1.19.2, ogb==1.3.1 and scikit-learn==0.24.0.

\subsection{Additional Experimental Results}
\label{addition}

As a supplement to Table \ref{table:semi}, we compare Meta-PN with the baseline methods on one more low-resource semi-supervised node classification task (10-shot) and the test results are summarized in Table \ref{table:10-shot}. Based on the results, we can observe that the proposed Meta-PN can significantly outperform all the baseline methods for the 10-shot task on different datasets, which further illustrates the effectiveness of Meta-PN for low-resource semi-supervised node classification.

    

\begin{table}[h!]
\centering
\caption{Test accuracy on 10-shot semi-supervised node classification with different models: Mean accuracy ($\%$) $\pm$ 95$\%$ confidence interval.}
\scalebox{0.82}{
\begin{tabular}{@{}lccccccccc@{}}
\toprule

\rule{0pt}{10pt} & \multicolumn{1}{c}{Cora-ML}  & \multicolumn{1}{c}{CiteSeer}  &  \multicolumn{1}{c}{PubMed} &  \multicolumn{1}{c}{MS-CS}

\\ \midrule

MLP       & $60.68\pm.48$   & $56.15\pm.69$   &  $63.97\pm.73$ & $85.67\pm.24$  \\
LP        & $72.38\pm.34$  & $59.97\pm.62$  &  $65.69\pm.98$ & $66.27\pm.22$  \\

GCN    & $78.48\pm.43$  & $69.04\pm.74$  &  $70.88\pm.75$ & $89.39\pm.19$  \\

SGC    & $74.40\pm.66$  & $72.79\pm.27$  &  $68.34\pm.96$ & $89.01\pm.31$    \\
\midrule

GLP   & $76.83\pm.25$  & $71.97\pm.16$  &  $73.65\pm.19$ & $90.11\pm.17$   \\
IGCN   & $78.32\pm.20$  & \underline{$73.32\pm.11$}  &  $74.45\pm.24$ & $88.26\pm.12$   \\
M3S   & $77.34\pm.25$  & $70.08\pm.19$  &  $72.78\pm.27$ & $89.01\pm.23$   \\
\midrule
APPNP   & \underline{$82.21\pm.24$}  & $72.70\pm.49$  &  $75.01\pm.67$ & \underline{$91.66\pm.11$}  \\
DAGNN   & $80.80\pm.31$  & $72.72\pm.35$  &  \underline{$76.70\pm.61$} & $91.60\pm.07$ \\

C\&S  & $75.38\pm.31$  & $69.61\pm.39$  &  $74.47\pm.18$ & $90.71\pm.29$ \\
GPR-GNN   & $78.82\pm.33$  & $69.83\pm.32$  &  $74.75\pm.26$ & $91.48\pm.14$ \\
\midrule

Meta-PN  & $\textbf{83.84}\pm\textbf{.29}$  & $\textbf{75.30}\pm\textbf{.42}$  &  $\textbf{78.44}\pm\textbf{.41}$ & $\textbf{92.26}\pm\textbf{.17}$ \\

\bottomrule
\end{tabular}}
\label{table:10-shot}

\end{table}

\begin{wrapfigure}{R}{0.5\textwidth} 
    \center
    \vspace{-0.4cm}
    \subfigure[APPNP]{
        \includegraphics[width=0.235\textwidth]{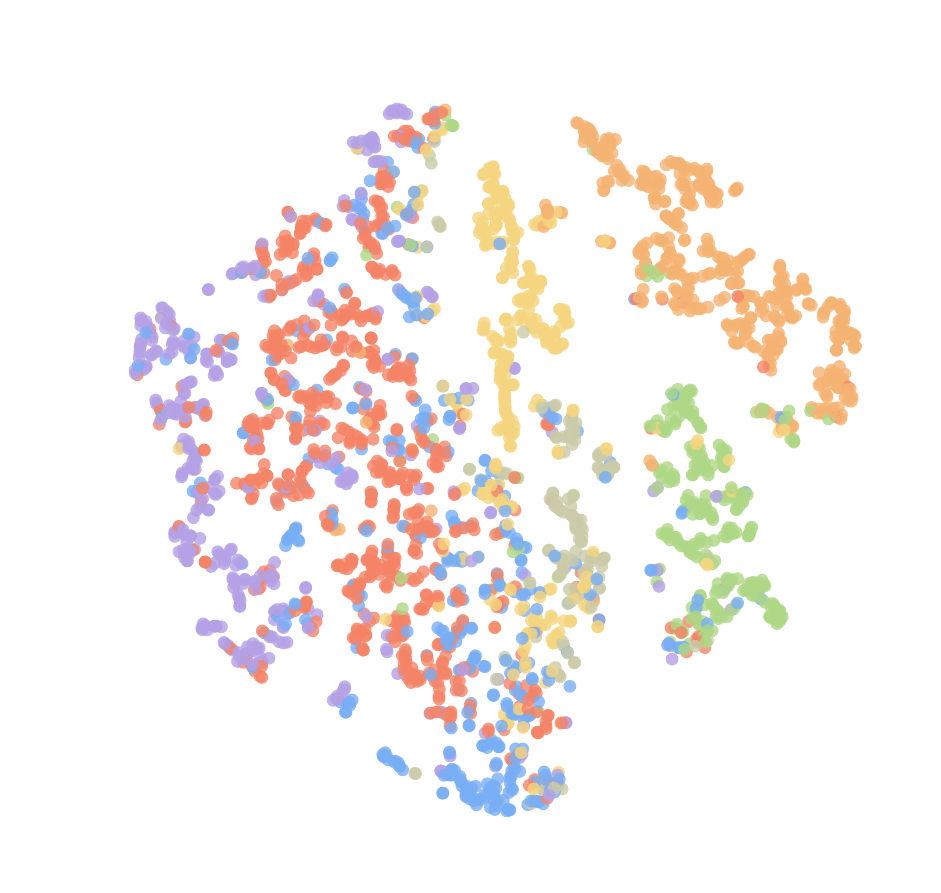}
    }
    \hspace{-0.2cm}
    \subfigure[Meta-PN]{
        \includegraphics[width=0.235\textwidth]{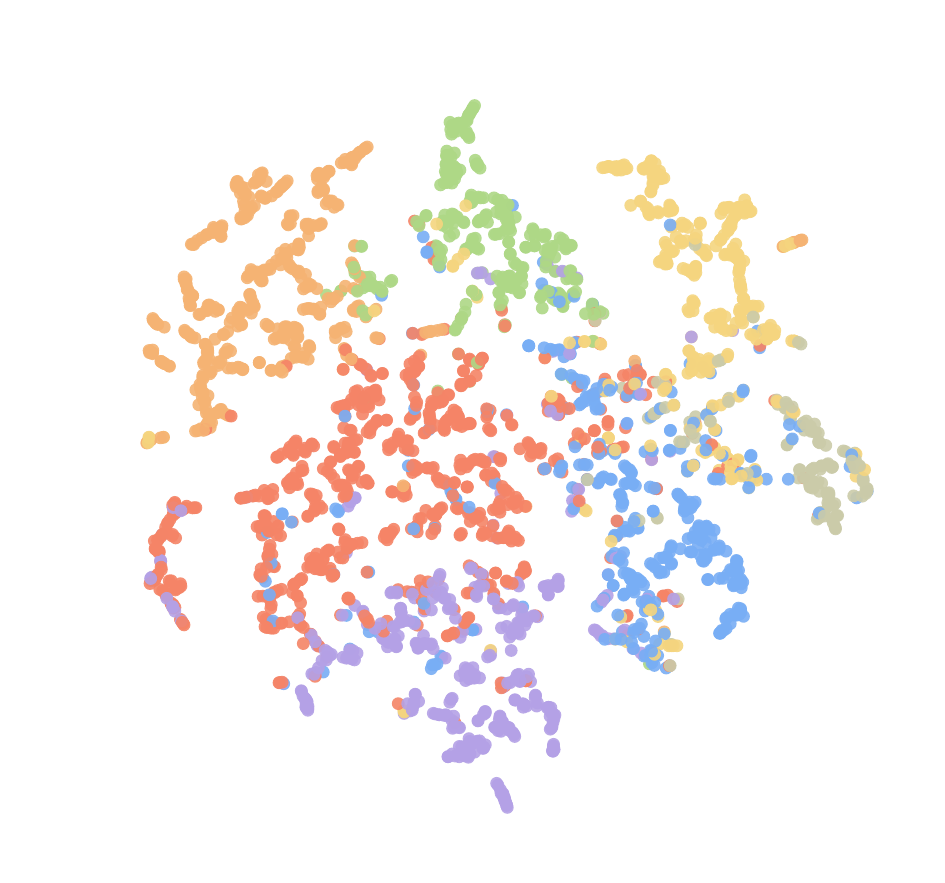}
    }
    \caption{Embedding visualization results of APPNP and Meta-PN on the Cora-ML dataset.}
    \vspace{-0.1cm}
    \label{fig:embed_vis}
\end{wrapfigure}

\smallskip
\noindent{\textbf{Embedding Visuliazation.}} To show the quality of the embedding from Meta-PN, we use t-SNE to visualize the extracted node representations from a strong baseline APPNP and Meta-PN for comparison. With the node's color denoting its label, from Figure \ref{fig:embed_vis} we can observe that though APPNP can effectively identify some of the classes, the boundary between different classes is still unclear. The proposed approach Meta-PN is able to generate more compact and separated clusters, which again verifies its superiority. 


\end{document}